\let\accentvec\vec
\documentclass[runningheads, envcountsame, a4paper]{llncs}

\let\vec\accentvec
\usepackage[pdftex]{graphicx}
\usepackage{epstopdf}
\usepackage[misc]{ifsym}
\usepackage[hyphens]{url}
\usepackage{microtype}

\usepackage{multirow}

\usepackage{amsthm,amsmath,amssymb}
\usepackage{mathrsfs}
\usepackage{bbm}
\usepackage{booktabs}
\usepackage{subfigure}
\usepackage[numbers,sort&compress,sectionbib]{natbib}

\begin{document}

\title{Scale Invariant Domain Generalization Image Recapture Detection}
\titlerunning{Scale Invariant Domain Generalization Image Recapture Detection}
\toctitle{Scale Invariant Domain Generalization Image Recapture Detection}

\author{Jinian Luo
\inst{1}
\and
Jie Guo\inst{2}$^{[\textrm{\Letter}]}$
\and
Weidong Qiu
\inst{2}
\and
Zheng Huang
\inst{2}
\and
Hong Hui
\inst{1}
}
\authorrunning{J. Luo et al.}

\institute{Institute of Cyber Science and Technology
\and
School of Cyber Science and Engineering\\ 
Shanghai Jiao Tong University, 800 Dongchuan Road, Shanghai 200240, P.R.China\\
\email{$\{$jinianluo,guojie,qiuwd,huang-zheng,huih$\}$@sjtu.edu.cn}}
\tocauthor{Jinian~Luo, Jie~Guo, Weidong~Qiu, Zheng~Huang, and Hong~Hui}

\maketitle
\setcounter{footnote}{0}

\begin{abstract}
    Recapturing and rebroadcasting of images are common attack methods in insurance frauds and face identification spoofing, and an increasing number of detection techniques were introduced to handle this problem. However, most of them ignored the domain generalization scenario and scale variances, with an inferior performance on domain shift situations, and normally were exacerbated by intra-domain and inter-domain scale variances. In this paper, we propose a scale alignment domain generalization framework (SADG) to address these challenges. First, an adversarial domain discriminator is exploited to minimize the discrepancies of image representation distributions among different domains. Meanwhile, we exploit triplet loss as a local constraint to achieve a clearer decision boundary. Moreover, a scale alignment loss is introduced as a global relationship regularization to force the image representations of the same class across different scales to be undistinguishable. Experimental results on four databases and comparison with state-of-the-art approaches show that better performance can be achieved using our framework.

\end{abstract}

\section{Introduction}
Digital images can now be easily obtained by cameras and distributed over the Internet. Modifications including recapturing are direct threats to image credibility at present. Thus, images as evidence require rigorous validation on the originality to be dependable testimonies~\cite{anjum2019recapture}.
According to ~\cite{cao2010identification}, human beings have difficulty discminating between recaptured and original images. To this end, recapture detection forensics are required to exclude recapture frauds.
\begin{figure}[htbp]  
    \centering  
    \includegraphics[width=\linewidth]{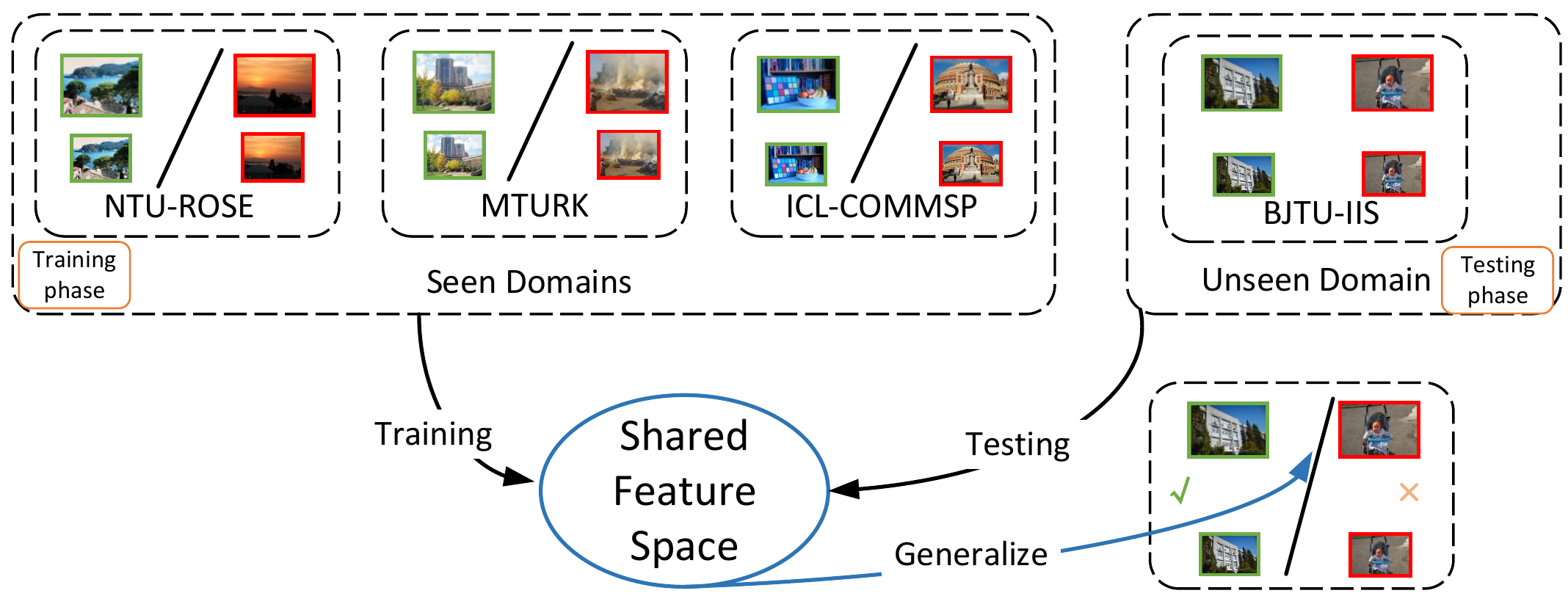}  
    \caption{Introduction to cross domain recapture detection. Our SADG method aims to learn a shared feature space which is robust with domain shift and scale variances scenarios. Images with green boarders are single captured and images with red boarders are recaptured. The upper row of images in each domain are of smaller scales and the lower row represents larger scales.}
    \label{introduction}
\end{figure}
In recapture detection tasks, wavelets statistical distributions ~\cite{farid2003higher, anjum2019recapture, sun2018recaptured}, texture distribution~\cite{cao2010identification} are exploited as detection features. According to ~\cite{li2017image}, texture distribution is considered to be a solid method in this task.
Physical traits, such as specularity, blurriness and chromaticity ~\cite{2015Li, gao2010single, lijian2017image} are also considered as an effective discriminative cue. DCT coefficients ~\cite{yang2017recaptured, yin2012markov}, quality ~\cite{zhu2018recaptured} and gray level co-occurance matrix (GLCM) ~\cite{awati2017classification} are also modeled. Besides, neural network methods ~\cite{yang2016recapture,li2017image,agarwal2018diverse, choi2017content} are proposed to further enhance the detection accuracy.
However, after compression on a recaptured image, the consequent deformation of texture feature patterns decreases the classification accuracy~\cite{2017Learning, gluckman2006scale, park2010multiresolution}. Besides, in order to build a robust recapture detection system, different datasets are collected for model training, which causes domain shift effects on properties of input images, including scale, illumination and color ~\cite{torralba2011unbiased}. Therefore, the distribution biases introduced by the dataset collection process is a practical challenge in recapture detection tasks. All of the above methods achieve successful performance only on single domain scenarios. Therefore, as is illustrated in Figure \ref{introduction}, cross domain recapture detection task is proposed to learn a shared feature space which preserves the most generalized classification cues and is robust with intra-domain and inter-domain scale variances.

Domain generalization (DG) methods are direct solutions for this task. Here a domain $\mathbb{D}$ = \{$\mathscr{X}$, $\mathbb{P}^{\textit{X}}$\} is defined by a feature space $\mathscr{X}$ and a marginal distribution $\mathbb{P}^{\textit{X}}$~\cite{pan2010survey}, and each single database is considered an independent domain in this paper. In a similar task, face-antispoofing, DG methods such as MADDG~\cite{shao2019multi} and SSDG~\cite{jia2020single} achieve promising performance in multi-domain scenarios. However, these DG methods are highly customized and consequently not applicable for recapture detection. Firstly, the recaptured images are only obtained from screens, thus recaptured features can be aggregated in feature space across different domains. Furthermore, the global relationship~\cite{2019Domain} such as scale variances can be exploited to enhance the discriminability of the representations. 

In this paper, as is illustrated in the left part of Figure \ref{introduction}, the shared feature space is learned from source domains. This paper makes the following contributions: (1) We introduce a competition between feature generator and domain discriminator for domain generalization; (2) 
In training phase the scale alignment(SA) operations are performed in each category across all source domains to aggregate the embedded features of different scale levels but the same capture category; (3) To improve the local representation compactness and further enhance the discriminability of generalized features, a triplet mining strategy is incorporated in the framework.
\section{Method}
\subsection{Overview}
This framework consists of three modules, i.e.(1) a domain discriminator competing  with feature generator; (2) a global scale alignment loss on the classification outputs of large and small scales, alongside the cross-entropy loss; (3) a triplet loss applied to the feature space as a local constraint. Details are described in the following sections.
\begin{figure*}[tb]  
    \centering  
    \includegraphics[width=0.95\textwidth]{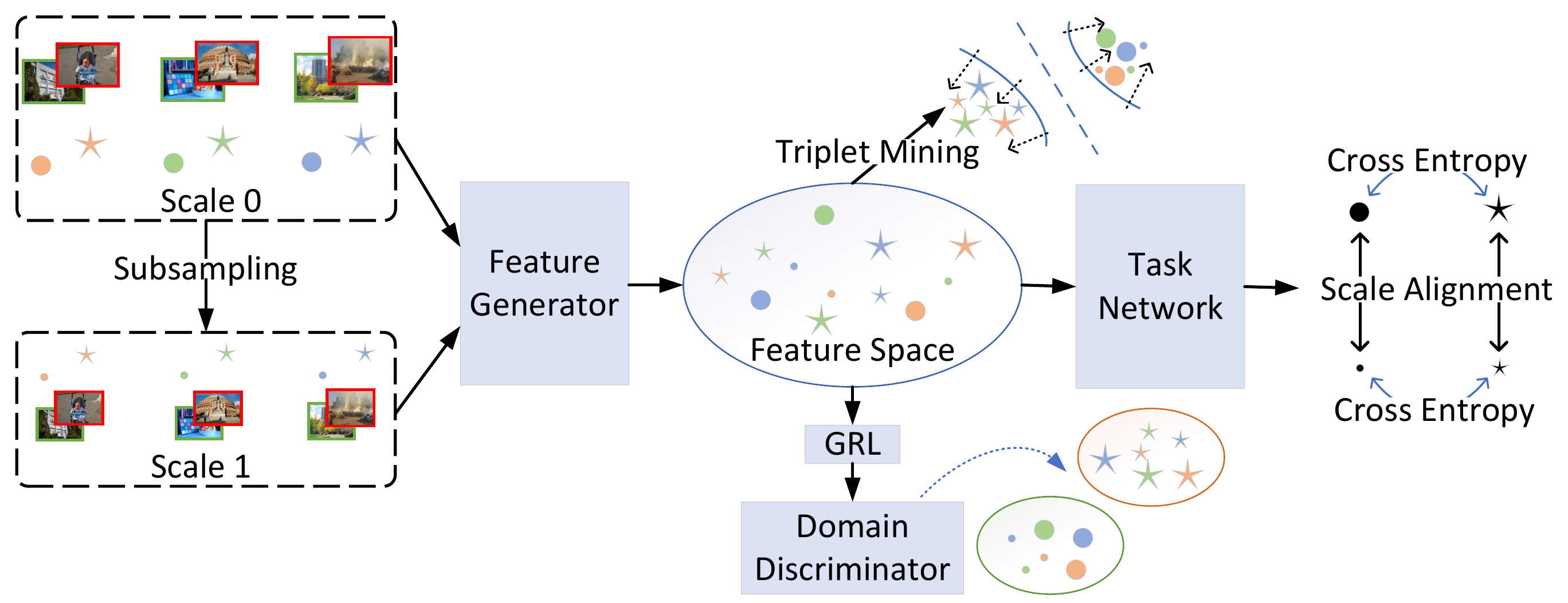}  
    \caption{Overview of the proposed method.
    Dots and stars represent single captured and recaptured images respectively. Symbols of three different colors represent different domains. Red border represents recaptured images, green border represents single captured images. Scale 1 represents larger scales and scale 0 represents smaller scales.}  
    \label{Prop_method}
\end{figure*}
\subsection{Adversarial Learning}
Suppose we have \textit{N} domains $\mathcal{D}$ = \{$\mathbb{D}_{1}$, $\mathbb{D}_{2}$, ..., $\mathbb{D}_{N}$\} and corresponding labels $\mathcal{Y}$ = \{$\mathbf{Y}_{1}$, $\mathbf{Y}_{2}$, ..., $\mathbf{Y}_{N}$\}. There are \textit{C} = 2 categories in each domain, where Y = 0/1 represents single capture/recapture. Our goal is to generalize from $\mathcal{D}$ and $\mathcal{Y}$ to unseen target domain $\mathbb{D}_{N+1}$. Here labels in target domain are not necessary for practical purposes. In recapture detection tasks, we postulate that common discriminative cues exist in both categories in sight of the identical nature of data collection in each domain and each class. To this end, we introduce adversarial learning method to the embedded feature space to exploit generalized differentiation information and minimize distribution bias of any specific source domain. 

For a feature generator $\textit{G}$, network input $\textbf{X}_{r}$ (recaptured images) and $\textbf{X}_{s}$ (single captured images) are transformed into embedded features $\textbf{F}_{r}$ and $\textbf{F}_{s}$:
\begin{equation}
    {\textbf{F}_r = \textit{G}(\textbf{X}_r),\  \textbf{F}_s = \textit{G}(\textbf{X}_s)} \label{feature_generator}
\end{equation}
A domain discriminator $\textit{D}$ is applied on $\textbf{F}_r$ and $\textbf{F}_s$ to determine their corresponding source domain:
\begin{equation}
    \textit{d}_r = \textit{D}(\textbf{F}_r),\  \textit{d}_s = \textit{D}(\textbf{F}_s) \quad  
    \label{domain_discriminator}
\end{equation}
There is a competition between domain discriminator \textit{D} and feature generator \textit{G}, where \textit{G} is trained to fool \textit{D} to make domain label indistinguishable from the shared discriminative feature space. Domain discriminator and feature generator are trained simultaneously and adversarially across all of source domains and categories in the training phase. Furthermore, in order to optimize domain discriminator and feature generator in the same backpropagation step, a gradient reverse layer(GRL)~\cite{jia2020single,ganin2015unsupervised} is inserted between them. The task of domain discriminator is effectively a multiclass classification, thus we utilize cross-entropy loss to measure the performance of \textit{G} and \textit{D}:
\begin{equation}
    \begin{split}
\mathop{\min}\limits_{\textit{D}} 
\mathop{\max}\limits_{\textit{G}} 
\mathcal{L}_{Ada}(\textit{X}, \textit{Y}_D; \textit{G}, \textit{D})
=\ -\mathbb{E}_{x, y \sim{\textit{X}, \textit{Y}_D}}
\sum_{n = 1}^{N}\mathbbm{1}[n=y]
\log \textit{D}(\textit{G}(x))\label{adversarial_loss}
    \end{split}
\end{equation}
where $\textit{Y}_D$ is a set of domain labels.
\subsection{Scale Alignment Clustering}
Images from different domains or even the same domain have different scales~\cite{2017Learning}, which adversely affects the generalization performance. Inspired by the global class alignment objective in MASF~\cite{2019Domain}, we propose to introduce a scale alignment objective to the distribution of classification outputs and structure the feature space by an explicit regularization. Our preliminary experiments demonstrated that scale relationship is better represented in the classification outputs than in the feature space, therefore, scale alignment is performed on task network outputs, which is different from \cite{2019Domain}. For each class $\textit{c}$, the concept of scale is modeled by $\textbf{s}_{c}^{(jl)}$ for large scale and $\textbf{s}_{c}^{(js)}$ for small scale:
\begin{equation}
    \textbf{s}_{c}^{(jl)}\ =\ \textit{T}(\textit{G}(\textbf{x}_{c}^{(jl)})),\  
    \textbf{s}_{c}^{(js)}\ =\ \textit{T}(\textit{G}(\textbf{x}_{c}^{(js)}))\label{scale_concept}
\end{equation}
where \textit{T} and \textit{G} represents task network and feature generator, and \textit{j} is the index of the image from the synthesized training dataset. We define scale alignment loss on the distribution of $\textbf{s}_{c}^{(jl)}$ and $\textbf{s}_{c}^{(js)}$:
\begin{equation}
    \begin{split}
    \mathop{\min}\limits_{\textit{G, T}}
    \mathcal{L}_{SA}(\textit{G}, \textit{T};\ \mathbb{D}_{r},\ \mathbb{D}_{s})
    =
    &\frac{1}{C}
    \sum_{c = 1}^{C}
    \frac{1}{N}
    \sum_{j = 1}^{N}
    \frac{1}{2}
    [D_{KL}(\textbf{s}_{c}^{(jl)}\  \| \ \textbf{s}_{c}^{(js)})\ \\
    &\ +\ D_{KL}(\textbf{s}_{c}^{(js)}\  \| \ \textbf{s}_{c}^{(jl)})]
    \label{scale_alignment_loss}
    \end{split}
\end{equation}
where $D_{KL}(\textbf{p}\|\textbf{q}) = \sum_{r} \textit{p}_{r}\log\frac{\textit{p}_{r}}{\textit{q}_{r}}$ and symmetric Kullback-Leibler (KL) divergence is $\frac{1}{2}[D_{KL}(\textbf{p}\|\textbf{q}) + D_{KL}(\textbf{q}\|\textbf{p})]$. The discrepency between distributions of $\textbf{s}_{c}^{(jl)}$ and $\textbf{s}_{c}^{(js)}$ are measured by symmetric KL divergence across two classes.
\subsection{Triplet Mining}
To hold local feature compactness~\cite{2019Domain} in the feature space, we insert a triplet loss to aggregate intra-class samples and separate inter-class samples for a clearer decision boundary. 
Triplet mining is also introduced by Jia et al.~\cite{jia2020single} and Shao et al.~\cite{shao2019multi} with a view to structure the feature space. However, in recapture detection context, images are recaptured only from screens, thus in contrast we enforce feature compactness in both classes regardless of domain or scale.
Specifically, we assume there are three source domains in training phase. In a triplet mining procedure, recaptured and single captured images are recollected from all source domains.

The two objectives in triplet mining are: pull apart recapture samples from single capture samples, and aggregate each class respectively. In the backpropagation step, feature generator \textit{G} is optimized by:
\begin{equation}
    \begin{split}
        &\mathop{\min}\limits_{\textit{G}}
        \quad \mathcal{L}_{DA\_Trip}(\textit{G};\  \textbf{X}_r,\  \textbf{X}_s) \ = \ \\
    &\sum_{\begin{array}{c}
        \forall y_a = y_p, y_a \ne y_n\\
        i = j
    \end{array}} 
    [\|\textit{G}(x_{i}^{a}) - \textit{G}(x_{j}^{p})\|_{2}^{2}\ -\  \|\textit{G}(x_{i}^{a}) - \textit{G}(x_{j}^{n})\|_{2}^{2}\ + \alpha]_{+} \\
    &+ \ \sum_{\begin{array}{c}
        \forall y_a = y_p, y_a \ne y_n\\
        i \ne j
    \end{array}} 
    [\|\textit{G}(x_{i}^{a}) - \textit{G}(x_{j}^{p})\|_{2}^{2}\ -\  \|\textit{G}(x_{i}^{a}) - \textit{G}(x_{j}^{n})\|_{2}^{2}\ + \alpha]_{+}
    \label{Triplet_equation}
\end{split}
\end{equation}
where \textit{G} denotes the feature generator, superscripts \textit{a} and \textit{n} represents different classes and \textit{a} and \textit{p} samples stem from the same class. Subscripts \textit{i} and \textit{j} indicates there is no restriction on domain or scale of samples. $\alpha$ represents a pre-defined positive parameter. Finally, samples from different domains or scales but the same category are forced to be more compact in the feature space.
\subsection{Scale Invariant Domain Generalization}
We formulate the integrated framework into an optimization objective as follow:
\begin{equation}
    \mathcal{L}_{SADG}
    =
    \mathcal{L}_{Cls}
    + \lambda_{1} \mathcal{L}_{Ada}
    + \lambda_{2} \mathcal{L}_{DA\_Trip}
    + \lambda_{3} \mathcal{L}_{SA}
    \label{finalobjective}
\end{equation}
where $\mathcal{L}_{Cls}$ is a task-specific loss function. Because recapture detection is a classification task, the framework is optimized by cross-entropy loss, denoted by $\mathcal{L}_{Cls}$. $\lambda_{1}$, $\lambda_{2}$ and $\lambda_{3}$ are pre-defined parameters to balance four losses. As is illustrated in Figure \ref{Prop_method}, this framework is trained in an end-to-end manner in the training phase. After training, \textit{G} achieves a more generalized feature space, which is robust with domain shift and scale variances.
\section{Experiments}
\subsection{Experimentsal Settings}
\subsubsection{Dataset} Four recapture detection datasets are collected to simulate real-life scenarios and evaluate our proposed method against other baseline methods. Our selected datasets are BJTU-IIS~\cite{2015Li} (B for short), ICL-COMMSP~\cite{thongkamwitoon2015image,muammar2013investigation} (I for short), mturk~\cite{agarwal2018diverse} (M for short) and NTU-ROSE~\cite{cao2010identification,hong2011statistical} (N for short), and the number of recapture/single capture images and recapture devices are shown in Table \ref{tabledataset}.
\begin{table}[htbp]
    \centering
    \caption{Four experimental datasets.}
    \label{tabledataset}
    \resizebox{\linewidth}{!}{
        \begin{tabular}{c|c|c|c}
            \toprule
            \textbf{Dataset} & \textbf{Recapture Device Count} & \textbf{Recaptured Image Count} & \textbf{Single Captured Image Count} \\
            \hline
            B & 2 & 706 & 636 \\
            I & 8 & 1440 & 905 \\
            M & 119 & 1369 & 1368 \\
            N & 9 & 2776 & 2712 \\
            \bottomrule
        \end{tabular}
    }
\end{table}
These datasets are collected for different purposes, specifically, B is for evaluation on high resolution images; images in I are controlled by distance between camera and screen; images in N are captured in a lighting controlled room; and for M, the aim was to crowd-source the collection of images. The contents, illumination, scales and resolution of pictures are different across datasets or within a single dataset.
\subsubsection{Implementation Details} Our work is implemented using Pytorch as a framework. The images are cropped  $256 \times 256 \times 3$ off, in a RGB color space. ResNet-18 is exploited as backbone of feature generator. In order to achieve a better generalization performance, the Adam optimizer learning rate is set to 1e-4, and batch size is 8 for each domain. So the total batch size is 24 in a 3 source domains case, and 16 in a limited source experiment. We set the hyperparameters $\lambda_{1}$, $\lambda_{2}$ and $\lambda_{3}$ in Equation \ref{finalobjective} to be 0.1, 0.2 and 0.1, respectively. Each time one domain is chosen as target, and remaining three domains are source domains. Thus, there are four experimental tasks in total.
\subsection{Experimental Comparison}
\begin{figure*}[htbp]
  \centering
  \resizebox{\textwidth}{!}{
  \subfigure{
  \includegraphics[]{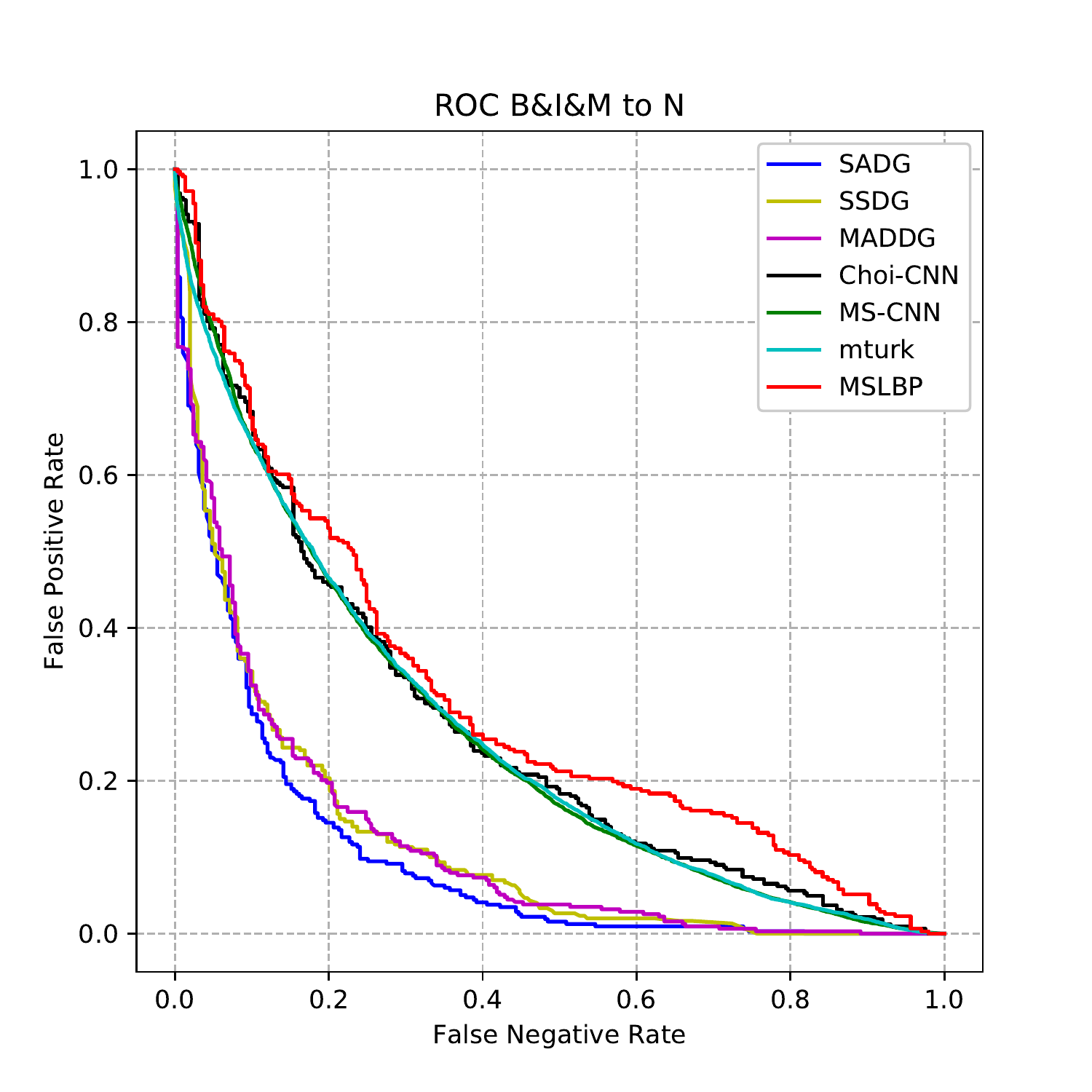}
  }
  \subfigure{
  \includegraphics[]{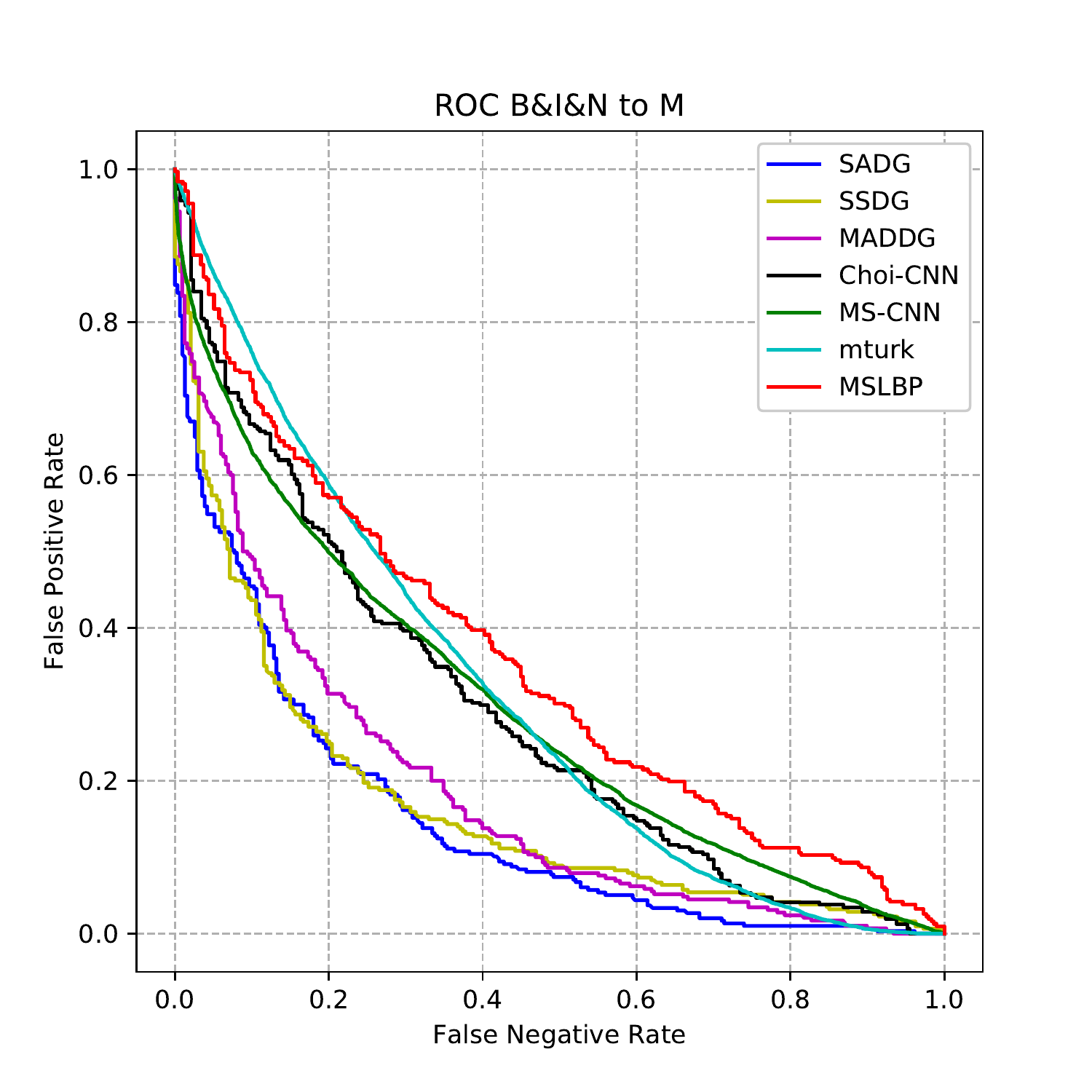}
  }}
  \caption{ROC curves for two recapture detection domain generalization experiments.}
  \label{roc}
\end{figure*}
\subsubsection{Baseline Methods}
We compare our proposed SADG framework with several state-of-the-art recapture detection methods, multi-scale learning methods and domain generalization algorithms: \textbf{Multi-scale LBP}(\textbf{MS-LBP})~\cite{cao2010identification}, \textbf{Choi-CNN}~\cite{choi2017content}, \textbf{Multi-scale CNN} (\textbf{MS-CNN})~\cite{2017Learning}, \textbf{mturk}~\cite{agarwal2018diverse}, \textbf{MADDG}~\cite{shao2019multi}, \textbf{SSDG}~\cite{jia2020single}. The MS-LBP and MS-CNN are multi-scale methods, Choi-CNN and mturk are deep learning methods for recapture detection. MADDG and SSDG are two algorithms for face anti-spoofing task, and we compare these two methods with SADG because no recapture detection methods pay attention to domain generalization to our best knowledge.
\subsubsection{Comparison Results}
\begin{figure*}[htbp]
    \centering
    \resizebox{\textwidth}{!}{
    \subfigure{
    \includegraphics[]{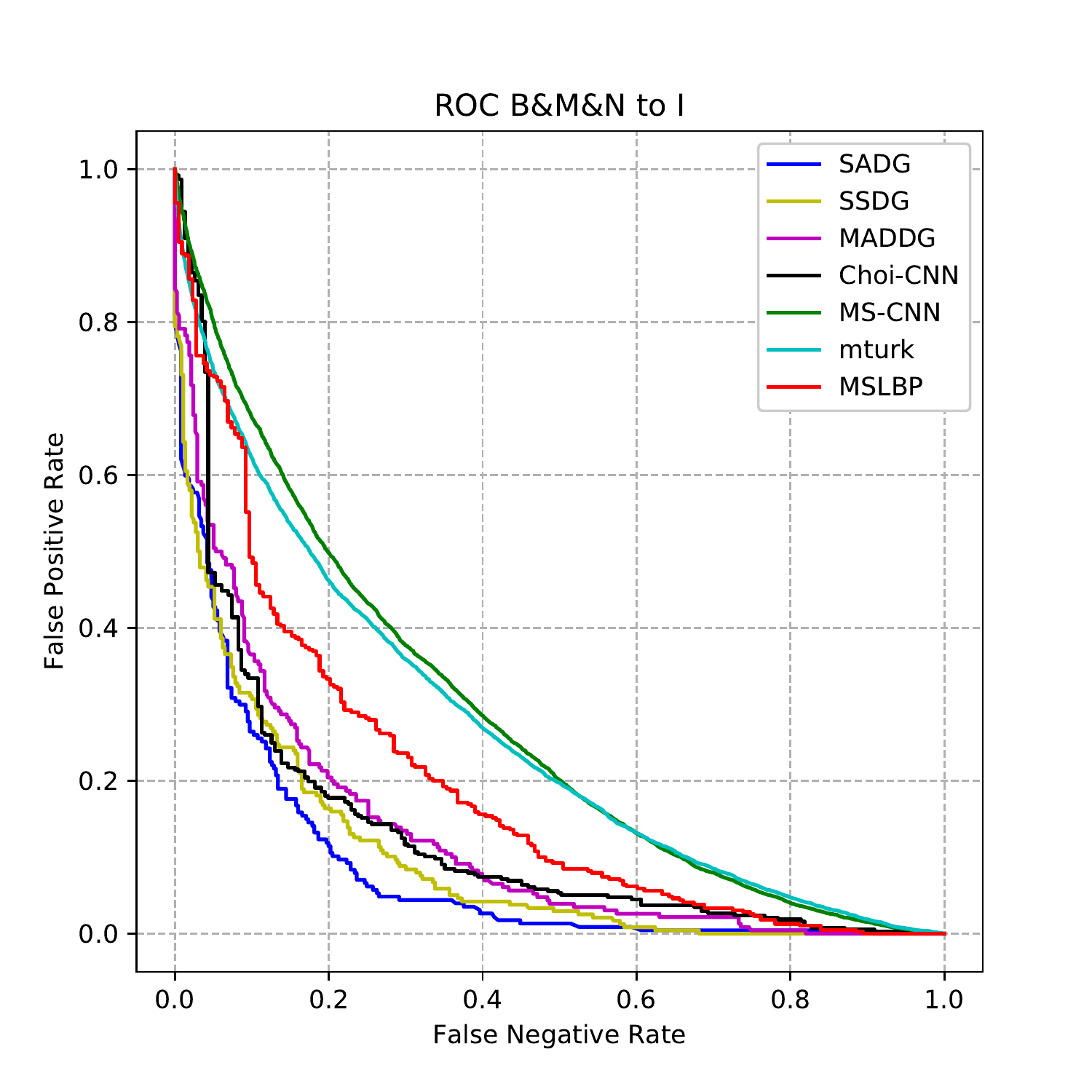}
    }
    \subfigure{
    \includegraphics[]{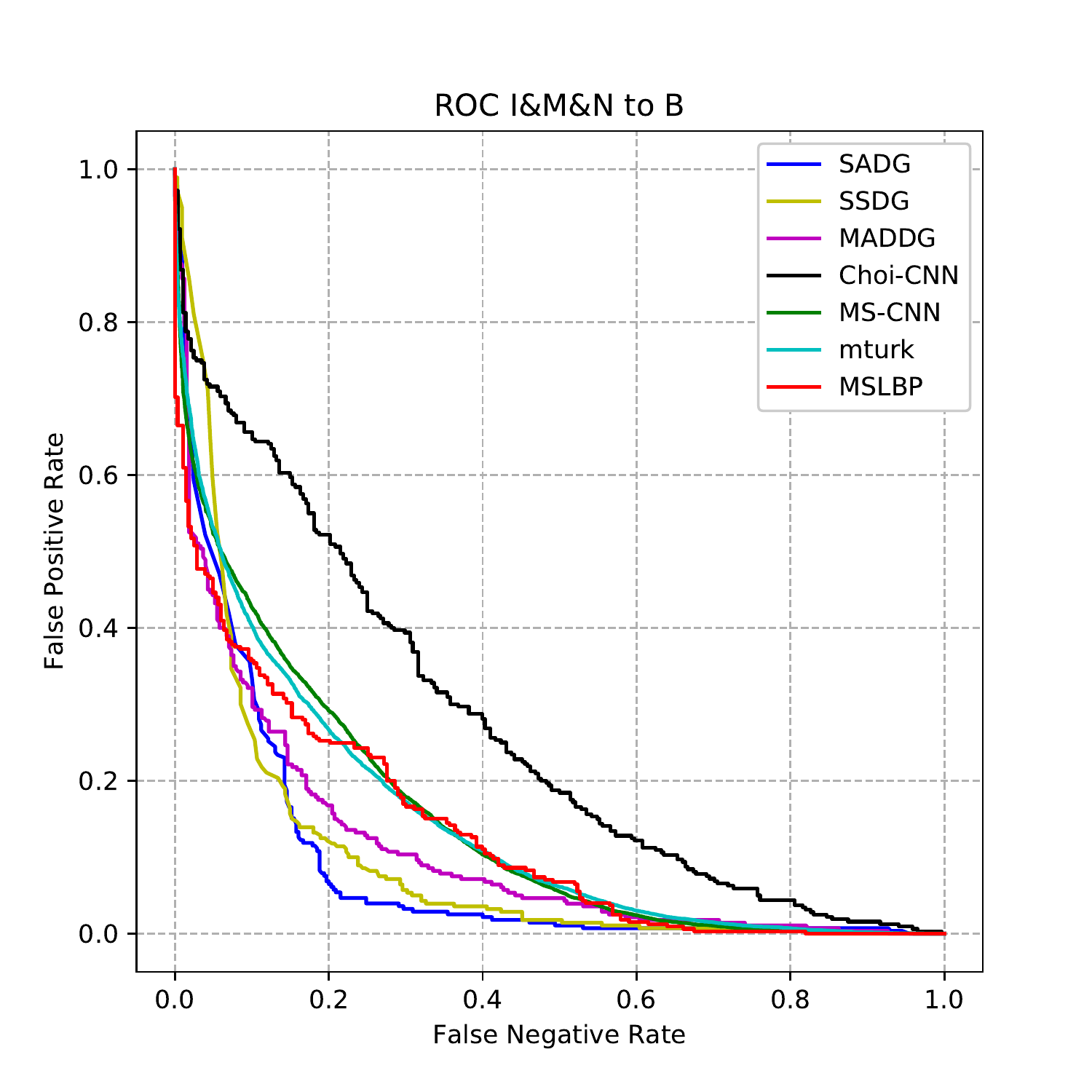}
    }}
    \caption{ROC curves for two recapture detection domain generalization experiments.}
    \label{rocb}
  \end{figure*}

  As is shown in Table \ref{table_SOTA} and \ref{table_review}, our algorithm outperforms all compared state-of-the-art methods except the HTER in experiment I\&M\&N to B, with only 0.01\% behind SSDG.  By subsampling, the scale variances are amplified in target domains. From these two tables, we can see that scale variances are crucial to detection performance. In the second column of Table \ref{tabledataset}, the recapture device counts are significantly different. Furthermore, the variation of scale and resolution are larger in M than in B, and thus all of the methods performs better in experiment I\&M\&N to B than in experiment B\&I\&N to M, but SADG is better than all other methods, which demonstrates the effectiveness of our scale alignment loss and domain generalization strategy. As is shown in Figure \ref{roc} and Figure \ref{rocb}, when compared with traditional methods, the proposed method performs better because other methods pay no attention on domain shift and cannot achieve a generalized feature space. Moreover, although SSDG and MADDG are domain generalization methods, neither of them focuses on the scale variances, either within a single domain or among different domains. The domain shift affects introduced by scale variances can be addressed by adversarial learning, but scale variances also exists in a single domain. This was resolved by global alignment and clustering operations of SADG in a pairwise manner.
\begin{table*}[htbp]
    \centering
    \caption{Comparison with state-of-the-art methods on four recapture detection domain generalization experiments with original and subsampled scales.}
    \resizebox{\textwidth}{!}{
    \begin{tabular}{c|c|c|c|c|c|c|c|c}
        \toprule
        \multirow{2}{*}{\textbf{Method}} & \multicolumn{2}{c|}{\textbf{B\&I\&M to N}} & \multicolumn{2}{c|}{\textbf{B\&I\&N to M}} & \multicolumn{2}{c|}{\textbf{B\&M\&N to I}} & \multicolumn{2}{c}{\textbf{I\&M\&N to B}} \\
        \cline{2-9}
        \multirow{2}{*}{} & HTER(\%) & AUC(\%) & HTER(\%) & AUC(\%) & HTER(\%) & AUC(\%) & HTER(\%) & AUC(\%) \\
        \hline
        \hline
        MS-LBP~\cite{cao2010identification} & 33.07 & 69.85 & 39.50 & 65.09 & 35.86 & 81.08 & 20.56 & 87.25\\
        Choi-CNN~\cite{choi2017content} & 24.87 & 73.60 & 47.70 & 71.60 & 37.99 & 87.20 & 30.39 & 73.47\\
        MS-CNN~\cite{2017Learning} & 32.50 & 74.43 & 28.59 & 70.90 & 38.91 & 72.35 & 15.00 & 85.72\\
        mturk~\cite{agarwal2018diverse} & 32.81 & 74.41 & 35.16 & 69.32 & 36.88 & 73.72 & 18.25 & 85.88\\
        MADDG~\cite{shao2019multi} & 19.74 & 88.39 & 26.15 & 81.72 & 20.40 & 87.64 & 18.25 & 89.40\\
        SSDG~\cite{jia2020single} & 20.06 & 88.41 & 22.37 & 83.67 & 18.43 & 90.59 & \textbf{15.12} & 90.48\\
        \hline
        Ours(\textbf{SADG}) & \textbf{15.95} & \textbf{90.28} & \textbf{22.20} & \textbf{85.60} & \textbf{15.93} & \textbf{92.03} & 15.13 & \textbf{91.65}\\
        \bottomrule
    \end{tabular}
    }
    \label{table_SOTA}
  \end{table*}
\begin{table*}[htbp]
    \centering
    \caption{Comparison with state-of-the-art methods on domain generalization experiments with original scale.}
    \resizebox{\textwidth}{!}{
    \begin{tabular}{c|c|c|c|c|c|c|c|c}
        \toprule
        \multirow{2}{*}{\textbf{Method}} & \multicolumn{2}{c|}{\textbf{B\&I\&M to N}} & \multicolumn{2}{c|}{\textbf{B\&I\&N to M}} & \multicolumn{2}{c|}{\textbf{B\&M\&N to I}} & \multicolumn{2}{c}{\textbf{I\&M\&N to B}} \\
        \cline{2-9}
        \multirow{2}{*}{} & HTER(\%) & AUC(\%) & HTER(\%) & AUC(\%) & HTER(\%) & AUC(\%) & HTER(\%) & AUC(\%) \\
        \hline
        \hline
        MADDG~\cite{shao2019multi} & 18.75 & 90.06 & 10.85 & 96.09 & 14.64 & 93.32 & 12.99 & 94.17\\
        SSDG~\cite{jia2020single} & 16.61 & 91.58 & 12.66 & 94.93 & 12.17 & 93.84 & 11.02 & 94.85\\
        \hline
        Ours(\textbf{SADG}) & \textbf{16.45} & \textbf{91.42} & \textbf{10.03} & \textbf{96.14} & \textbf{11.18} & \textbf{95.28} & \textbf{8.55} & \textbf{94.85}\\
        \bottomrule
    \end{tabular}
    }
    \label{table_review}
  \end{table*}
\subsection{Discussion}
\subsubsection{Ablation Study}
\begin{table*}[htbp]
  \centering
  \caption{Ablation study on the effectiveness of three components in the framework.}
  \label{tabel3}
  \resizebox{\textwidth}{!}{
  \begin{tabular}{c|c|c|c|c|c|c|c|c}
      \toprule
      \multirow{2}{*}{\textbf{Method}} & \multicolumn{2}{c|}{\textbf{B\&I\&M to N}} & \multicolumn{2}{c|}{\textbf{B\&I\&N to M}} & \multicolumn{2}{c|}{\textbf{B\&M\&N to I}} & \multicolumn{2}{c}{\textbf{I\&M\&N to B}} \\
      \cline{2-9}
      \multirow{2}{*}{} & HTER(\%) & AUC(\%) & HTER(\%) & AUC(\%) & HTER(\%) & AUC(\%) & HTER(\%) & AUC(\%) \\
      \hline
      \hline
      SADG wo/ad & 21.05 & 87.73 & 25.49 & 82.16 & 23.87 & 83.23 & 15.46 & \textbf{92.66}\\
      SADG wo/trip & 24.18 & 83.17 & 25.16 & 82.92 & 23.84 & 85.58 & 16.77 & 91.63\\
      SADG wo/sa & 18.75 & 88.27 & 23.25 & 84.15 & 17.59 & 90.69 & 15.46 & 91.88\\
      \hline
      Ours(\textbf{SADG}) & \textbf{15.95} & \textbf{90.28} & \textbf{22.20} & \textbf{85.60} & \textbf{15.93} & \textbf{92.03} & \textbf{15.13} & 91.65\\
      \bottomrule
  \end{tabular}
  }
  \label{table_ABLATION}
\end{table*}

Because every component, adversarial learning, triplet mining and scale alignment clustering in SADG framework are independent from domain settings, we conduct ablation study on the aforementioned four sets of domain generalization experiments, eliminating effects of one component each time. \textbf{SADG wo/ad} denotes the SADG framework without adversarial learning, where we disengage GRL and domain discriminator from the backpropagation procedure. This specific network does not explicitly exploit the shared information in feature space. \textbf{SADG wo/trip} denotes the SADG framework without triplet mining, where the effects of triplet loss are canceled. In this case, the framework does not utilize local clustering objective as a regularization. \textbf{SADG wo/sa} denotes the SADG framework without scale alignment clustering, where the global relationship alignment between different scales are removed from feature space. 

Table \ref{table_ABLATION} shows the performance of every incomplete SADG framework degrades on each set of domain generalization experiments. As expected, this result verifies that each component in SADG advances the performance simultaneously by global and local alignment and clustering operations, and that the intact version of proposed scheme achieves the finest performance.
\subsubsection{Stages Comparison}
\begin{table*}[htbp]
    \centering
    \caption{Comparison with other scale alignment strategies on domain generalization.}
    \label{tabel2}
    \resizebox{\textwidth}{!}{
    \begin{tabular}{c|c|c|c|c|c|c|c|c}
        \toprule
        \multirow{2}{*}{\textbf{Method}} & \multicolumn{2}{c|}{\textbf{B\&I\&M to N}} & \multicolumn{2}{c|}{\textbf{B\&I\&N to M}} & \multicolumn{2}{c|}{\textbf{B\&M\&N to I}} & \multicolumn{2}{c}{\textbf{I\&M\&N to B}} \\
        \cline{2-9}
        \multirow{2}{*}{} & HTER(\%) & AUC(\%) & HTER(\%) & AUC(\%) & HTER(\%) & AUC(\%) & HTER(\%) & AUC(\%) \\
        \hline
        \hline
        feature-SADG & 18.43 & 85.98 & 23.68 & 83.16 & 16.93 & 91.44 & \textbf{15.12} & \textbf{91.87}\\
        task-SADG & 16.94 & 87.76 & 23.85 & 83.37 & 16.45 & 91.15 & 17.43 & 87.39\\
        \hline
        Ours(\textbf{SADG}) & \textbf{15.95} & \textbf{90.28} & \textbf{22.20} & \textbf{85.60} & \textbf{15.93} & \textbf{92.03} & 15.13 & 91.65\\
        \bottomrule
    \end{tabular}
    }
    \label{table_STAGE}
\end{table*}
According to MASF~\cite{2019Domain}, the concept of a class is represented by the average of embedded features in the feature space. However, our preliminary experiments indicated that KL divergence output alignment strategy is better in scale alignment. There is also an average strategy deployed on task network classification score. We conduct an extensive study on these different strategies. The first strategy is a feature scale alignment. Following the work of~\cite{2019Domain}, the large (\textit{l}) scale concept of class \textit{c} is:
\begin{equation}
    \bar{\textbf{z}}_{c}^{(l)}
    =
    \frac{1}{\textit{N}_{l}^{(c)}}
    \sum_{n:y_{n}^{(l)} = c}
    \textit{G}(\textit{x}_{n}^{(l)})
    \label{concept}
\end{equation}
where \textit{G} is the feature generator. And we can define soft label distribution and loss as such:
\begin{equation}
    \textit{s}_{c}^{(l)}
    =
    softmax(\textit{T}(\bar{\textbf{z}}_{c}^{(l)}))
    \label{softmaxfeature}
\end{equation}
\begin{equation}
    \mathcal{L}_{SA}
    =
    \frac{1}{C}
    \sum_{c=1}^{C}
    \frac{1}{2}
    [ 
        D_{KL}(\textit{s}_{c}^{(l)} \| \textit{s}_{c}^{(s)})
        +
        D_{KL}(\textit{s}_{c}^{(s)} \| \textit{s}_{c}^{(l)})
    ]
    \label{safeature}
\end{equation}
where \textit{l} and \textit{s} indicate large scale and small scale respectively. The second strategy is a task network classification score alignment. The average classification score of large scale and scale alignment loss as such:
\begin{equation}
    \bar{s}_{c}^{(l)}
    =
    \frac{1}{N}
    \sum_{j=1}^{N}
    \textit{s}_{c}^{(jl)}
\end{equation}
\begin{equation}
    \mathcal{L}_{SA}^{\prime}
    =
    \frac{1}{C}
    \sum_{c=1}^{C}
    \frac{1}{2}
    [ 
        D_{KL}(\bar{\textit{s}}_{c}^{(l)} \| \bar{\textit{s}}_{c}^{(s)})
        +
        D_{KL}(\bar{\textit{s}}_{c}^{(s)} \| \bar{\textit{s}}_{c}^{(l)})
    ]
\end{equation}
The third strategy is the KL divergence output alignment described in Equations \ref{scale_concept} and \ref{scale_alignment_loss}.
Here the KL divergence is calculated in a pairwise manner. 

Table \ref{table_STAGE} shows that the last strategy outperforms the other two strategies except for the I\&M\&N to B experiment, where the proposed method is 0.01\% behind the feature scale alignment method. Therefore, our KL divergence alignment strategy is the robustest method.
\subsubsection{Limited Source Domains}
\begin{table}[htbp]
    \centering
    \caption{Comparison with state-of-the-art methods on limited source domains for recapture detection.}
    \begin{tabular}{c|c|c|c|c}
        \toprule
        \multirow{2}{*}{\textbf{Method}} & \multicolumn{2}{c|}{\textbf{M\&N to B}} & \multicolumn{2}{c}{\textbf{M\&N to I}} \\
        \cline{2-5}
        \multirow{2}{*}{} & HTER(\%) & AUC(\%) & HTER(\%) & AUC(\%)\\
        \hline
        \hline
        MS-LBP~\cite{cao2010identification} & 24.83 & 83.15 & 36.33 & 81.58\\
        Choi-CNN~\cite{choi2017content} & 45.70 & 81.30 & 48.05 & 72.24\\
        MS-CNN~\cite{2017Learning} & 24.34 & 83.48 & 30.13 & 76.95\\
        SSDG~\cite{jia2020single} & 22.03 & 85.38 & 23.93 & 84.08\\
        \hline
        Ours(\textbf{SADG}) & \textbf{18.56} & \textbf{88.81} & \textbf{22.94} & \textbf{84.31}\\
        \bottomrule
    \end{tabular}
    
    \label{table_LIMITED}
\end{table}
We further conduct experimental comparison in a limited source scenario (e.g. only two source domains are available), which is a normal case in real-life practices. The scale variance between mturk (M for short) and NTU-ROSE (N for short) is more significant than that of the other two domains, thus we choose M and N as source domains and the remaining two as target domains. In Table \ref{table_LIMITED}, our proposed method outperforms other methods significantly. When compared with Table \ref{table_SOTA}, our proposed method performs better by expoliting the scale variance information. Therefore, our method achieves more generalized feature space even in a extremely limited source scenario.
\subsection{Conclusion}
To address two challenges, scale variances and domain shift, we propose a scale alignment domain generalization framework (SADG). Different from existing recapture detection methods, our SADG framework exploits generalized discriminative information in shared feature space. Moreover, we apply global and local regularization on the embedded features. Specifically, the global relationship between different scales is aligned and utilized for optimization. Meanwhile, triplet loss is also incorporated as a further constraint for class clustering and a clearer decision boundary. Extensive experiments on public databases validate the effectiveness of our proposed method and prove that our SADG framework achieves state-of-the-art results in domain generalization recapture detection.
\section*{Acknowledgements}
Portions of the research in this paper used the ROSE Recaptured Image Dataset made available by the ROSE Lab at the Nanyang Technological University, Singapore.


\begin{thebibliography}{99}
    
  \bibitem{agarwal2018diverse}
  Agarwal, S., Fan, W., Farid, H.: A diverse large-scale dataset for evaluating
    rebroadcast attacks. In: 2018 IEEE International Conference on Acoustics,
    Speech and Signal Processing (ICASSP). pp. 1997--2001. IEEE (2018)
  
  \bibitem{anjum2019recapture}
  Anjum, A., Islam, S.: Recapture detection technique based on edge-types by
    analysing high-frequency components in digital images acquired through lcd
    screens. Multimedia Tools and Applications pp. 1--21 (2019)
  
  \bibitem{awati2017classification}
  Awati, C., Alzende, N.H.: Classification of singly captured and recaptured
    images using sparse dictionaries. International Journal  \textbf{5}(7) (2017)
  
  \bibitem{cao2010identification}
  Cao, H., Kot, A.C.: Identification of recaptured photographs on lcd screens.
    In: 2010 IEEE International Conference on Acoustics, Speech and Signal
    Processing. pp. 1790--1793. IEEE (2010)
  
  \bibitem{choi2017content}
  Choi, H.Y., Jang, H.U., Son, J., Kim, D., Lee, H.K.: Content recapture
    detection based on convolutional neural networks. In: International
    Conference on Information Science and Applications. pp. 339--346. Springer
    (2017)
  
  \bibitem{2019Domain}
  Dou, Q., Castro, D.C., Kamnitsas, K., Glocker, B.: Domain generalization via
    model-agnostic learning of semantic features. In: Advances in Neural
    Information Processing Systems (NeurIPS). vol.~32. Vancouver, BC, Canada
    (2019)
  
  \bibitem{farid2003higher}
  Farid, H., Lyu, S.: Higher-order wavelet statistics and their application to
    digital forensics. In: 2003 Conference on computer vision and pattern
    recognition workshop. vol.~8, pp. 94--94. IEEE (2003)
  
  \bibitem{ganin2015unsupervised}
  Ganin, Y., Lempitsky, V.: Unsupervised domain adaptation by backpropagation.
    In: International conference on machine learning. pp. 1180--1189. PMLR (2015)
  
  \bibitem{gao2010single}
  Gao, X., Ng, T.T., Qiu, B., Chang, S.F.: Single-view recaptured image detection
    based on physics-based features. In: 2010 IEEE International Conference on
    Multimedia and Expo. pp. 1469--1474. IEEE (2010)
  
  \bibitem{gluckman2006scale}
  Gluckman, J.: Scale variant image pyramids. In: 2006 IEEE Computer Society
    Conference on Computer Vision and Pattern Recognition (CVPR'06). vol.~1, pp.
    1069--1075. IEEE (2006)
  
  \bibitem{hong2011statistical}
  Hong, C.: Statistical image source model identification and forgery detection.
    Ph.D. thesis, Nanyang Technological University (2011)
  
  \bibitem{jia2020single}
  Jia, Y., Zhang, J., Shan, S., Chen, X.: Single-side domain generalization for
    face anti-spoofing. In: Proceedings of the IEEE/CVF Conference on Computer
    Vision and Pattern Recognition. pp. 8484--8493 (2020)
  
  \bibitem{li2017image}
  Li, H., Wang, S., Kot, A.C.: Image recapture detection with convolutional and
    recurrent neural networks. Electronic Imaging  \textbf{2017}(7),  87--91
    (2017)
  
  \bibitem{lijian2017image}
  Li, J., Wu, G.: Image recapture detection through residual-based local
    descriptors and machine learning. In: International Conference on Cloud
    Computing and Security. pp. 653--660. Springer (2017)
  
  \bibitem{2015Li}
  Li, R., Ni, R., Zhao, Y.: An effective detection method based on physical
    traits of recaptured images on lcd screens. In: International Workshop on
    Digital Watermarking. pp. 107--116. Springer (2015)
  
  \bibitem{muammar2013investigation}
  Muammar, H., Dragotti, P.L.: An investigation into aliasing in images
    recaptured from an lcd monitor using a digital camera. In: 2013 IEEE
    International Conference on Acoustics, Speech and Signal Processing. pp.
    2242--2246. IEEE (2013)
  
  \bibitem{2017Learning}
  Noord, N.V., Postma, E.: Learning scale-variant and scale-invariant features
    for deep image classification. Pattern Recognition  \textbf{61},  583--592
    (2017)
  
  \bibitem{pan2010survey}
  Pan, S.J., Yang, Q.: A survey on transfer learning ieee transactions on
    knowledge and data engineering. 22 (10): 1345  \textbf{1359} (2010)
  
  \bibitem{park2010multiresolution}
  Park, D., Ramanan, D., Fowlkes, C.: Multiresolution models for object
    detection. In: European conference on computer vision. pp. 241--254. Springer
    (2010)
  
  \bibitem{shao2019multi}
  Shao, R., Lan, X., Li, J., Yuen, P.C.: Multi-adversarial discriminative deep
    domain generalization for face presentation attack detection. In: Proceedings
    of the IEEE/CVF Conference on Computer Vision and Pattern Recognition. pp.
    10023--10031 (2019)
  
  \bibitem{sun2018recaptured}
  Sun, Y., Shen, X., Lv, Y., Liu, C.: Recaptured image forensics algorithm based
    on multi-resolution wavelet transformation and noise analysis. International
    Journal of Pattern Recognition and Artificial Intelligence  \textbf{32}(02),
    1854003 (2018)
  
  \bibitem{thongkamwitoon2015image}
  Thongkamwitoon, T., Muammar, H., Dragotti, P.L.: An image recapture detection
    algorithm based on learning dictionaries of edge profiles. IEEE Transactions
    on Information Forensics and Security  \textbf{10}(5),  953--968 (2015)
  
  \bibitem{torralba2011unbiased}
  Torralba, A., Efros, A.A.: Unbiased look at dataset bias. In: CVPR 2011. pp.
    1521--1528. IEEE (2011)
  
  \bibitem{yang2017recaptured}
  Yang, P., Li, R., Ni, R., Zhao, Y.: Recaptured image forensics based on quality
    aware and histogram feature. In: International Workshop on Digital
    Watermarking. pp. 31--41. Springer (2017)
  
  \bibitem{yang2016recapture}
  Yang, P., Ni, R., Zhao, Y.: Recapture image forensics based on laplacian
    convolutional neural networks. In: International Workshop on Digital
    Watermarking. pp. 119--128. Springer (2016)
  
  \bibitem{yin2012markov}
  Yin, J., Fang, Y.: Markov-based image forensics for photographic copying from
    printed picture. In: Proceedings of the 20th ACM international conference on
    Multimedia. pp. 1113--1116 (2012)
  
  \bibitem{zhu2018recaptured}
  Zhu, N., Li, Z.: Recaptured image detection through enhanced residual-based
    correlation coefficients. In: International Conference on Cloud Computing and
    Security. pp. 624--634. Springer (2018)
    
\end{thebibliography}
\end{document}